\documentstyle{article}

\newcommand{\ie}{{\em i.e.\ }}
\newcommand{\eg}{{\em e.g.\ }}
\newcommand{\ra}{$\rightarrow$}

\author{Andrew Fordham \\ SCS Research Group \\ Department of
  Sociology \\ University of Surrey \\ Guildford \\ Surrey, GU2 5XH,
  UK \\{\tt ajf@soc.surrey.ac.uk}\and
  Matthew Crocker \\ Centre for Cognitive Science \\ University of
  Edinburgh \\ 2 Buccleuch Place \\ Edinburgh, EH8 9LW \\ Scotland \\
  {\tt mwc@cogsci.ed.ac.uk}}
\title{Parsing with Principles and Probabilities\footnote{\copyright
    1994 Association for Computational Linguistics. This paper appears
    in the Proceedings of the Balancing Act Workshop: Combining
    Symbolic and Statistical Approaches to Language, 1st June 1994,
    Las Cruces, New Mexico.} \\ {\normalsize cmp-lg/9408004}}

\begin{document}
\maketitle

\bibliographystyle{abbrv}

\begin{abstract}
  This paper is an attempt to bring together two approaches to
  language analysis. The possible use of probabilistic
  information in principle-based grammars and parsers is considered,
  including discussion on some theoretical and computational problems
  that arise. Finally a partial implementation of these ideas is
  presented, along with some preliminary results from testing on a
  small set of sentences.
\end{abstract}

\section{Introduction}
\label{sec:intro}

Both principle-based parsing and probabilistic methods for the
analysis of natural language have become popular in the last decade.
While the former borrows from advanced linguistic specifications of
syntax, the latter has been more concerned
with extracting distributional regularities from language to aid the
implementation of NLP systems and the analysis of corpora.

These symbolic and statistical approaches are
beginning to draw together as
it becomes clear that one cannot exist entirely without the other: the
knowledge of language posited over the years by theoretical linguists
has been useful in constraining and guiding statistical approaches,
and the corpora now available to linguists have resurrected the desire
to account for real language data in a more principled way than had
previously been attempted.

This paper falls directly between these approaches,
using statistical information derived from corpora analysis to weight
syntactic analyses produced by a `principles and
parameters' parser. The use of probabilistic
information in principle-based grammars and parsers is considered,
including discussion on some theoretical and computational problems
that arise. Finally a partial implementation of these ideas is
presented, along with some preliminary results from testing on a
small set of sentences.

\section{Government-Binding Theory}
\label{sec:gbt}

\begin{figure}[htbp]
  \begin{center}
    \leavevmode
\setlength{\unitlength}{0.0125in}%
\begin{picture}(260,202)(50,610)
\thicklines
\put(160,705){\line( 3,-4){ 49.200}}
\put(145,705){\line(-3,-4){ 49.200}}
\put(120,800){\makebox(0,0)[lb]{\raisebox{0pt}[0pt][0pt]{\bf D-Structure}}}
\put(120,710){\makebox(0,0)[lb]{\raisebox{0pt}[0pt][0pt]{\bf S-Structure}}}
\put( 50,625){\makebox(0,0)[lb]{\raisebox{0pt}[0pt][0pt]{\bf Phonetic Form}}}
\put(175,625){\makebox(0,0)[lb]{\raisebox{0pt}[0pt][0pt]{\bf Logical Form}}}
\put(155,795){\line( 0,-1){ 70}}
\put( 90,760){\makebox(0,0)[lb]{\raisebox{0pt}[0pt][0pt]{\it   move-$\alpha$}}}
\put(280,785){\makebox(0,0)[lb]{\raisebox{0pt}[0pt][0pt]{\sf
$\theta$-criterion)}}}
\put( 95,670){\makebox(0,0)[lb]{\raisebox{0pt}[0pt][0pt]{\it pf}}}
\put(195,670){\makebox(0,0)[lb]{\raisebox{0pt}[0pt][0pt]{\it lf-movement
(move-$\alpha$)}}}
\put(205,710){\makebox(0,0)[lb]{\raisebox{0pt}[0pt][0pt]{\sf (Case Theory,
Subjacency)}}}
\put(310,610){\makebox(0,0)[lb]{\raisebox{0pt}[0pt][0pt]{\sf Binding Theory)}}}
\put(270,625){\makebox(0,0)[lb]{\raisebox{0pt}[0pt][0pt]{\sf (Empty Category
Principle, }}}
\put(205,800){\makebox(0,0)[lb]{\raisebox{0pt}[0pt][0pt]{\sf ($\bar{{\rm
X}}$-theory, lexical insertion, }}}
\end{picture}

 \end{center}
  \caption{The T-model of grammar}
  \label{fig:tmodel}
\end{figure}
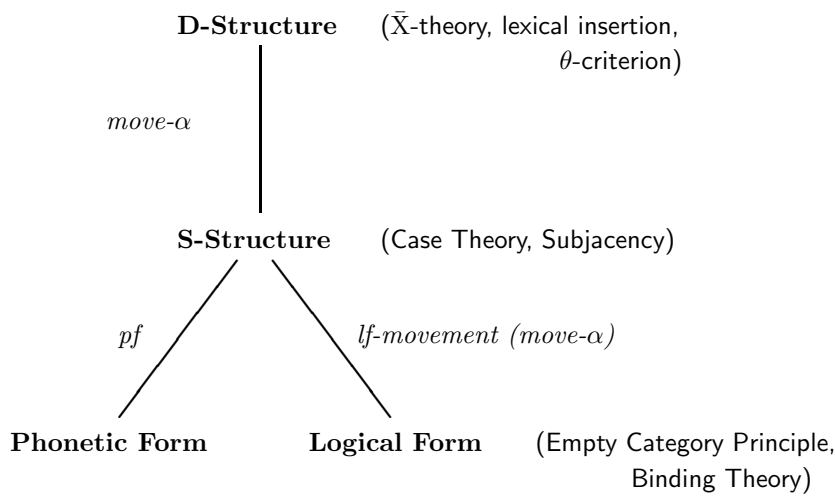

The principles and parameters paradigm in linguistics is most fully
realised in the Government-Binding Theory (GB) of Chomsky
\cite{chomsky81,chomsky86} and others. The grammar is divided
into modules which filter out ungrammatical structures at the various
levels of representation; these levels are related by general
transformations. A sketch of the organisation of GB (the `T-model') is
shown in figure \ref{fig:tmodel}.

Little work has been done on the complexity of algorithms
used to parse with a principle-based grammar, since such grammars do
not exist as accepted mathematically well-defined constructs. It has
been estimated that in general, principle-based parsing
can only be accomplished in exponential time, \ie
$O(2^n)$ \cite{berwick84,weinberg88}.

A feature of principle-based grammars is their potential to
assign some meaningful representation to a string which is strictly
ungrammatical. It is an inherent feature of phrase structure grammars
that they classify the strings of words from a language into two
(infinite) sets, one containing the grammatical strings and the other
the {\bf un}grammatical strings. Although attempts have been made to
modify PS grammars/parsers to cope with extragrammatical
input, \eg
\cite{carbonell83,douglas92,jensen83,mellish89}, this is a feature
which has to be `added on' and tends to affect the statement of the
grammar.

Due to the lack of an accepted formalism for the specification
of principle-based grammars, Crocker and Lewin \cite{crocker92}
define the declarative
`Proper Branch' formalism, which can be used with a number of
different parsing methods.

A proper branch is a set of three nodes --- a mother and two daughters
--- which are constructed by the parser, using a simple mechanism such
as a shift-reduce interpreter, and then `licensed' by the principles
of grammar. A complete phrase marker of the input string can then be
constructed by following the manner in which the mother node from one
proper branch is used as a daughter node in a dominating proper
branch.

Each proper branch is a binary branching structure, and so all
grammatical constraints will need to be encoded locally. Crocker
\cite{crockerphd} develops ``a
`representational' reformulation of the transformational model which
decomposes syntactic analysis into several representation types ---
including phrase structure, chains, and coindexation --- allowing one
to maintain the strictly local characterisation of principles with
respect to their relevant representation types,''
\cite[p.~511]{crocker92}.

By using the proper branch method of axiomatising the grammar, the
structure building section of the parser is only constrained in that
it must produce proper branches; it is therefore possible to
experiment with different interpreters (\ie structure proposing
engines) while keeping the grammar constant.

\section{The Grammar and Parser}
\label{sec:gp}

A small principle-based parser was built, following
the proper branch formalism developed in \cite{crocker92}.
Although the grammar is very limited, the use of probabilities in
ranking the
parser's output can be seen as a first step towards implementing a
principle-based parser using a more fully specified collection of
grammar modules.

The grammar is
loosely based on three modules taken from Government-Binding Theory
--- X-bar theory, Theta Theory and Case
Theory. Although these embody the spirit of the constraints found in
Chomsky \cite{chomsky81} they are not intended to be entirely
faithful to this specification of syntactic theory. There is also only
a single level of representation (which is explicitly constructed for
output purposes but not consulted by the parser). This representation
is interpreted as S-structure.

Explanations of the knowledge contained within each grammar principle
is given in the following sections.

\subsection{\={X} Theory}
\label{sec:xbar}

X-bar Theory uses a set of schemata to license local subtrees. We
use a parametrised version of the X-bar schemata, similar
to that of Muysken \cite{muysken83}, but employing features which
relate to the state of the head word's theta grid to give five
schemata (figure \ref{fig:xbar})
\begin{figure}[htbp]
  \begin{center}
    \leavevmode
\begin{minipage}{1.75in}
    \begin{enumerate}
\item X$^{-}_{-}$ \ra Y$^{-}_{-}$ X$^+_{-}$

\item X$^{S}_{+}$ \ra X$^{S}_{+}$ Y$^+_{-}$

\item X$^{S}_{-}$ \ra X$^{S}_{+}$ Y$^-_{-}$

\item X$^{S}_{-}$ \ra X$^{S}_{-}$ Y$^-_{-}$

\item X$^{S}_{-}$ \ra Y$^{-}_{-}$ X$^S_{-}$
\end{enumerate}
\end{minipage}
  \end{center}
  \caption{The X-bar Schemata}
  \label{fig:xbar}
\end{figure}.
A node includes the following features (among others):
\begin{enumerate}
\item Category: the standard category names are employed.
\item Specifier ({\sf SPEC}): this feature specifies whether the word
  at the head of the phrase being built requires a specifier.
\item Complement ({\sf COMP}): the complement feature is redundant in
  that the information used to derive it's value is already present in
  a word's theta grid, and will therefore be checked for well-formedness
  by the theta criterion. Since
  this information is not referenced until later, the {\sf COMP} feature is
  used to limit the number of superfluous proper-branches generated by
  the parser.
\item The head (\ie lexical item) of a node is carried on each
  projection of that node along with its theta grid.
\end{enumerate}

The probabilities for occurrences of the X-bar schema
were obtained from sentences from the preliminary Penn Treebank
corpus of the Wall Street Journal, chosen
because of their length and the head of their verb phrase (\ie the
main verbs were all from the set for which theta role data was
obtained); the examples were manually parsed by the authors.

The probabilities were calculated using the
following equation, where $X^{S_i}_{C_i} \rightarrow Y^{S_j}_{C_j}\;
Z^{S_k}_{C_k}$ is a specific schema, $\bar{X}$ is the set of X-bar
schemata and $A$ and $B$ and $C$ are variables over category, {\sf
  SPEC} and {\sf COMP} feature bundles:
\begin{equation}
  P(X^{S_i}_{C_i} \rightarrow Y^{S_j}_{C_j}\; Z^{S_k}_{C_k} | \bar{X}) =
  \frac{C(X^{S_i}_{C_i} \rightarrow Y^{S_j}_{C_j}\;
    Z^{S_k}_{C_k})}{C(A \rightarrow B\; C)}
\end{equation}

This is different to manner in which probabilities are collected for
stochastic context-free grammars, where the identity of the mother node is
taken into account, as in the equation below:
\begin{equation}
  P(X^{S_i}_{C_i} \rightarrow Y^{S_j}_{C_j}\; Z^{S_k}_{C_k} | \bar{X}) =
  \frac{C(X^{S_i}_{C_i} \rightarrow Y^{S_j}_{C_j}\;
    Z^{S_k}_{C_k})}{C(X^{S_i}_{C_i} \rightarrow B\; C)}
\end{equation}
This would result in misleading probabilities for the X-bar schemata
since the use of schemata (3), (4), and (5)
would immediately bring down the probability of a parse compared to
a parse of the same string which happened to use only (1) and
(2).\footnote{The probabilities for (1) and (2) would be~1 as they
  have unique mothers.}

The overall
(X-bar) likelihood of a parse can then be computed by multiplying
together all the probabilities obtained from each application of the
schemata, in a manner analogous to that used to obtain the
probability of a phrase marker generated by an SCFG.
Using the schemata in this way suggests that the building of structure
is category independent, \ie it is just as likely that a verb will
have a (filled) specifier position as it is for a noun. The work on
stochastic context-free grammars suggests a different set of results,
in that the specific categories involved in expansions are all
important.
While SCFGs will tend to deny that all categories expand in certain
ways with the same probabilities, they make this claim while using a
homogeneous grammar formalism. When a more modular theory is
employed, the source of the supposedly category specific information
is not as obvious. The use of lexical probabilities on specifier and
complement co-occurrence with specific heads (\ie lexical items) could
exihibit properties that appear to be category specific, but are in
fact caused by common properties which are shared by lexical items of
the same category.\footnote{It is of course possible to store these
  cross-item similarities as lexical rules \cite{bresnan78}, but this
  alone does not entail that the properties are specific to a
  category, {\em cf.\/}~the theta grids of verbs and their `related'
  nouns.} Since it can be argued that the probabilistic information
on lexical items  will be needed independently, there is no need to
use category specific information
in assigning probabilities to syntactic configurations.

\subsection{Theta Theory}
\label{sec:theta}

Theta theory is concerned with the assignment of an argument structure
to a sentence. A verb has a number of the thematic (or `theta')
roles which must be assigned to its arguments, \eg a transitive verb
has one theta role to `discharge' which must be assigned to an NP.

If a binary branching formalism is employed, or indeed any
formalism where the arguments of an item and the item itself are not
necessarily all sisters, the problem of when to access the
probability of a theta application is presented. The easiest method
of obtaining and applying theta probabilities will be
with reference to whole theta grids. Each theta grid for a word
will be assigned a probability which is not dependent on any
particular items in the grid, but rather on the occurrence of the
theta grid as a whole.

A preliminary version of
the Penn Treebank bracketed corpus was analysed to extract
information on the sisters of particular verbs. Although the Penn Treebank
data is unreliable since it does not always distinguish complements
from adjuncts, it was the only suitable parsed corpus to which the
authors had access. Although the distinction between complements and
adjuncts is a theoretically interesting one, the process of
determining which constructions fill which functional roles in the
analysis of real text often creates a number of problems (see
\cite{hindle93b} for discussion on this issue regarding output of the
Fidditch parser \cite{hindle93}).

The probabilities for each of the verbs' theta grids were calculated
using the equation below, where $P(s_i|v)$ is the probability of the
theta grid $s_i$ occurring with the verb $v$, $(v, s_i)$ is an
occurrence of the items in $s_i$ being licensed by $v$, and
$S$ ranges over all theta grids for $v$:
\begin{equation}
  P(s_i|v) = \frac{C(v, s_i)}{C(v, S)}
\end{equation}

\subsection{Case Theory}
\label{sec:pct}

In its simplest form, Case theory invokes the Case filter to ensure
that all noun phrases in a parse are assigned (abstract) case.
Case theory differs from both X-bar and Theta theory in that it is
category specific: only NPs require, or indeed can be assigned,
abstract case. If we are to implement a probabilistic version of a
modular grammar theory incorporating a Case component, a relevant
question is: are there multiple ways of assigning Case to  noun
phrases in a sentence? \ie can ambiguity arise due to the presence of
two candidate Case assigners?

Case theory suggests that the answer to this is negative, since Case
assignment is linked to theta theory {\em via\/} visibility, and it is
not possible for an NP to receive more than one theta role. As a
result, the use of Case probabilities in a parser would be at best
unimportant, since some form of ambiguity is needed in the module, \ie
it is possible to satisfy the Case filter in more than one way, for
probabilities associated with the module to be of any use. While
having a provision for using probabilities deduced from Case
information, the implemented parser does not in fact use Case in
its parse ranking operations.

\subsection{Local Calculation}
\label{sec:local}

The use of a heterogeneous grammar formalism and multiple
probabilities invokes the problem of their combination.
There are  at least two ways in which each mother's probabilities
can be calculated; firstly, the probability information of the same
type can be used: the daughters' X-bar probabilities alone could be
used in calculating the mother's X-bar probability. Alternatively, a
combination of
some or all of the daughters' probability features could be employed,
thus making, {\em e.g.\/}, the X-bar probability of the mother
dependent upon {\bf all} the stochastic information from the
daughters, including theta and Case probabilities, {\em etc\/}.

The need for a method of combining the daughter
probabilities into a useful figure for the calculation of the mother
probabilities is likely to involve trial and error, since theory thus
far has had nothing to say on the subject. The former method, using
only the relevant daughter probabilities, therefore seems to be the
most fruitful path to follow at the outset, since it does not require
a way of integrating probabilities from different modules while the
parse is in progress, nor is it as computationally expensive.

\subsection{Global Calculation}

The manner in which the global probability is calculated will
be partly dependent upon the information contained in the local
probability calculations.

If the probabilities for partial analyses have been calculated using
only probabilities of the same types from the subanalyses --- \eg X-bar,
Theta --- the probabilities at the top level will have been calculated
using informationally distinct figures. This has
the advantage of making `pure' probabilities available, in that the
X-bar probability will reflect the likelihood of the structure alone,
and will be `uncontaminated' by any other information. It should then
be possible to experiment with different methods of combining these
probabilities, other than the obvious `multiplying them together'
techniques, which could result in one type of probabililty emerging as
the most important.

On the other hand, if probabilities calculated during the parse take
all the different types of probabilities into account at each
calculation --- \ie the X-bar, theta, {\em etc.\/}~probabilities on
daughters are all taken into account when calculating the mother's
X-bar probability --- the probabilities at the top level will not be
pure, and a lot of the information contained in them will be redundant
since they will share a large subset of the probabilities used in
their separate calculations. It will not therefore be easy to gain
theoretical insight using these statistics, and their most profitable
method of combination is likely to be more haphazard affair than when
more pure probabilities are used.

The parser used in testing employed the first method and therefore
produced separate module probabilities for each node. For the lack of
a better, theoretically motivated method for combining these figures,
the product of the probabilities was taken
as the global probability for each parse.

\section{Testing the Parser}
\label{sec:test}

The parser was tested using sixteen sentences containing
verbs for which data had been collected from the Penn Treebank corpus.
The sentences were created by the authors to exhibit at least a degree
of ambiguity when it came to attaching a post-verbal phrase as an
adjunct or a complement. In order to force the choice of the `best'
parse on to the verb, the probabilities of theta grids for nouns,
prepositions, {\em etc.\/}~was kept
constant.

Of these 16 highest ranked parses, 7 are the expected parse, with
the other 9 exhibiting some form of mis-attachment. The fact that each
string received multiple
parses (the mean number of analyses being 9.135, and the median, 6)
suggests that the
probabilistic information did favourably guide the selection of a
single analysis.

It is not really possible to say from these results how successful the
whole approach of probabilistic principle-based parsing would be if it
were fully implemented. The inconclusive nature of the results
obtained was due to a number of limiting factors of the
implementation including the simplicity of the grammar and the lack of
available data.

\section{Discussion}
\label{sec:conc}

\subsection*{Limitations of the Grammar}
\label{sec:limgram}

The grammar employed is a partial
characterisation of Chomsky's Government-Binding theory
\cite{chomsky81,chomsky86} and only takes account of very local
constraints (\ie X-bar,
Theta and Case); a way of encoding all constraints in the
proper branch formalism (\eg \cite{crockerphd}) will be needed before a
grammar of sufficient coverage to be useful in corpora analysis can be
formulated. The
problem with using results obtained from the implementation given here
is that the grammar is sufficiently underspecified and so leaves too
great a task for the probabilistic information.

This approach could be viewed as putting the cart before the horse;
the usefulness of stochastic information in parsers presumes that a
certain level of accuracy can be achieved by the grammar alone. While
GB is an elegant theory of cognitive syntax, it has yet to be shown
that such a modular characteristion can be successfully employed in
corpus analysis.

\subsection*{Statistical Data and their Source}
\label{sec:stats}

The use of the preliminary Penn Treebank corpus for the extraction of
probabilities used in the implementation above was a choice forced by
lack of suitable materials. There are still very few parsed corpora
available, and none that contain information which is specified to the
level required by, {\em e.g.,\/} a GB grammar. While this is not an
absolute limitation, in that it is theoretically possible to extract
this information manually or semi-automatically from a corpus, time
constraints entailed the rejection of this approach.

It would be ultimately desirable if the use of probabilities in
principle-based parsing could be used to mirror the way that a
syntactic theory such as Government-Binding handles constructions ---
various modules of the grammar conspire to rule out illegal structures
or derivations. It would be an elegant result if a construction such
as the passive were to use probabilities for chains, Case assignment
{\em etc.\/} to select a parse that reflected the lexical changes that
had been undergone, \eg the greater likelihood of an NP featuring in
the verb's theta grid. It is this property of a number of modules
working hand in hand that needs to be carried over into the
probabilistic domain.

The objections that linguists once held against statistical methods are
disappearing slowly, partly due to results in corpora analysis that
show the inadequacy of linguistic theory when applied to naturally
occurring data. It is also the case that the rise of the connectionist
phoenix has brought the idea of weighted (though not strictly
probabilistic) functions of cognition back to the fore, freeing the
hands of linguists who believe that while an explanatorily adequate
theory of grammar is an elegant construct, its human implementation,
and its usage in computational linguists may not be straight forward.
This paper has hopefully shown that an integration of
statistical methods and current linguistic theory is a goal worth
pursuing.

\end{document}